\documentclass{article}

\usepackage{microtype}
\usepackage{graphicx}
\usepackage{subfigure}
\usepackage{booktabs} 
\usepackage{tabularx}
\usepackage{floatrow}
\usepackage{adjustbox}
\usepackage{bbm}
\usepackage{xcolor}
\usepackage{multicol}
\usepackage{amsmath}
\usepackage{amsthm}
\usepackage{amsfonts}
\usepackage{amssymb}
\usepackage{dblfloatfix}

\DeclareMathOperator*{\argmin}{argmin}
\DeclareMathOperator*{\argmax}{argmax}
\DeclareMathOperator*{\Prob}{\mathbb{P}}
\usepackage{hyperref}

\usepackage[accepted]{icml2018}

\icmltitlerunning{Penalizing Unfairness in Binary Classification}

\begin{document}

\twocolumn[
\icmltitle{Penalizing Unfairness in Binary Classification}



\icmlsetsymbol{equal}{*}

\begin{icmlauthorlist}
\icmlauthor{Yahav Bechavod}{huji}
\icmlauthor{Katrina Ligett}{huji}
\end{icmlauthorlist}

\icmlaffiliation{huji}{School of Computer Science and Engineering, The Hebrew University, Jerusalem, Israel}

\icmlcorrespondingauthor{Yahav Bechavod}{yahav.bechavod@cs.huji.ac.il}
\icmlcorrespondingauthor{Katrina Ligett}{katrina@cs.huji.ac.il}

\icmlkeywords{Regularization, Fairness, Equalized Odds, Equal Opportunity, COMPAS}

\vskip 0.3in
]



\printAffiliationsAndNotice{}  

\begin{abstract}
We present a new approach for mitigating unfairness in learned classifiers. In particular, we focus on binary classification tasks over individuals from two populations, where, as our criterion for fairness, we wish to achieve similar false positive rates in both populations, and similar false negative rates in both populations. As a proof of concept, we implement our approach and empirically evaluate its ability to achieve both fairness and accuracy, using datasets from the fields of criminal risk assessment, credit, lending, and college admissions.
\end{abstract}

\section{Introduction}
As machine learning-based methods have become increasingly prevalent in decision-making processes
that crucially affect people's lives, accuracy is no longer the sole
measure of a learning algorithm's success. In settings such as loan approvals~\cite{tracking}, policing~\cite{stopandfrisk}, targeted advertisement~\cite{latanyasweeney}, college admissions, or criminal risk assessments~\cite{machinebias}, algorithmic fairness must be carefully taken into account in order to ensure the absence of discrimination~\cite{bigdatasdisparateimpact, whiteguyproblem}.

Concerns of unfairness in classification were at the center of a recent media stir regarding the potential hazards of computer algorithms for risk assessment in the criminal justice system~\cite{machinebias,senttoprison}. The COMPAS system~(\citeyear{compasguide}), developed by Northpointe, is a proprietary algorithm, widely used in the United States for risk assessment and recidivism prediction. 
At the center of the controversy was an investigative report by Angwin et al.~(\citeyear{machinebias}), who observed that although the COMPAS algorithm demonstrated similar accuracy on whites and blacks when used to label individuals as either high or low risk for recidivism, the {\em direction} of errors made on whites versus blacks was very different.
More specifically, the rate of individuals who were classified using the COMPAS algorithm to be ``high risk'' but who did not actually re-offend was almost twice as high for black individuals as for whites; among those who were classified as ``low risk'' and did actually re-offend, the rate was significantly higher for whites than it was for blacks~\cite{ppanalysis}.

At least theoretically, fairness could necessarily come at a very high cost to accuracy, but it is possible that the tension between fairness and accuracy is far less stark on real-world data. Despite this, to date, there have been only a handful of techniques for ensuring fairness in classification that have been proposed and tested empirically.

\paragraph{Contribution} Motivated by this pressing need, we propose a new, easy-to-use, general-purpose technique for mitigating unfairness in classification settings. The approach deepens our understanding of how fairness considerations can be incorporated directly into the learning process, as opposed to imposing fairness post hoc on an arbitrary, unfair, learned classifier. We validate the ability of our approach to achieve both fairness and high accuracy, implementing and testing it on multiple datasets pertaining to recidivism, credit, loan defaults, and law school admissions. We find that our approach empirically outperforms existing approaches, and that fairness is often achievable at nearly no cost to accuracy.


\section{Related Work}

Approaches to algorithmic fairness generally fall into two categories---situations where no ground truth is known (or perhaps the notion of ground truth is not well-defined), and settings where the algorithm has access to labeled examples on which to learn (perhaps from historical examples). In situations without access to ground truth, typical approaches to fairness include changing the data (e.g., to prevent the learner from having direct/indirect access to attributes that are considered sensitive)~\cite{fairrepresentations,certifying,wordembeddings}, or adapting the classifier (e.g., to treat similar people similarly)~\cite{fairnessthroughawareness, fairnessbandits, kamishima}. When ground truth information is available, we wish to prevent situations where the algorithm errs {\em in favor} of one group within the population. In the specific context of criminal risk assessments, Berk et al.~(\citeyear{stateoftheart}) give a thorough comparison of various fairness notions. 
 

Both Kleinberg et al.~(\citeyear{inherent}) and Chouldechova~(\citeyear{chouldechova}) show that fair classification entails unavoidable trade-offs, and that there are a number of reasonable desiderata (calibration, matching false positive rates (FPR) across populations, and matching false negative rates (FNR) across populations), that cannot, in general, be achieved simultaneously~\cite{biasinevitable}. Follow-up work by Pleiss et al.~(\citeyear{calibration}) shows that even when calibration is compatible with a generalization of FPR- and FNR-matching, any algorithm achieving both must is no better than randomizing a percentage of the predictions of an existing classifier; further investigation of calibration as a criterion for fairness can be found in H{\'{e}}bert-Johnson et al.~(\citeyear{calibrationformasses}).

There are also computational challenges to fairness. Woodworth et al.~(\citeyear{woodworth}) show that even in the restricted case of learning linear predictors, assuming a convex loss function, and demanding that only the {\em sign} of the predictor needs to be non-discriminatory, the problem of matching FPR and FNR requires exponential time to solve in the worst case. They also point out that for many distributions and hypothesis classes, there may not exist a non-constant, deterministic, perfectly fair predictor. 

Despite these theoretical challenges, learning fair classifiers remains an important, practical problem that must be addressed on real data---decisions must be taken, and trade-offs must be made. To this end, there have been a number of recent specific technical proposals for achieving algorithmic fairness. 
The fairness objective we study in this paper, that of matching false positive and false negative rates across populations in classification tasks, has in particular received substantial attention in the literature. 

Hardt et al.~(\citeyear{hardt}) propose a post hoc approach for learning such a fair classifier, probabilistically flipping some of the decisions of a given (unfair) trained classifier in order to match FPR and FNR across populations. Their approach yields a predictor which is not restricted to any hypothesis class, and that is a (possibly randomized) function of the original (non-fair) learned predictor and of the sensitive attribute (population membership). Although this is an elegant and appealing idea, the Hardt et al.~approach only guarantees optimality for a strictly convex loss function and an unconstrained hypothesis class~\cite{woodworth}. Follow-up work of Woodworth et al.~(\citeyear{woodworth}), shows that, in many cases, any such post hoc approach might result in a highly sub-optimal classifier. As Woodworth et al.~conclude, {\em post-processing} an unfair classifier is sometimes insufficient to achieve the best possible combination of fairness and accuracy; rather, in some cases, fairness considerations should be actively integrated into the learning process. 

Zafar et al.~(\citeyear{disparatemistreatment}) give one such approach to integrating FPR and FNR matching into learning. Their algorithm relaxes the (non-convex) fairness constraints into proxy conditions, each in the form of a convex-concave (or, difference of convex) function. They then heuristically solve~\cite{convexconcave} the resulting optimization problem for a convex loss function.

The approach of the present work is to incorporate a penalty for unfairness into the learning objective. This is inspired in part by Kamishima et al.~(\citeyear{kamishima}), who designed an unfairness penalty term based on a very different notion of fairness, referred to in their paper as \textit{indirect prejudice}, which restricts the amount of mutual information between the prediction and the sensitive attribute. 

The present work introduces new penalty terms, designed to enforce matching of FPR and FNR. Our approach is easy to use, and general in the sense it can be plugged in and utilized in a range of learning settings concerning classification problems. The accuracy-fairness trade-offs of our approach empirically compare favorably with the algorithms of Zafar et al.~(\citeyear{disparatemistreatment}) and Hardt et al.~(\citeyear{hardt}) on the COMPAS dataset, and we further validate the performance of our approach on several additional datasets from other fields of interest.
\section{Fair Learning}
\subsection{Classical Approach}
In classical machine learning theory, when considering a classification task, the objective is typically to minimize a loss function that reflects the errors the chosen classifier makes on a fresh sample of data. One might naturally adjust the loss function to penalize differently for different sorts of errors (false positive or false negative, in the binary case), however, a priori, the classical approach does not do anything to control the distribution of errors across different sub-populations.

\subsection{Preliminaries}
We now introduce notation we will use to formalize our fairness objectives. We will represent each data point (person) as a pair $(x, y) \in \mathbb{R}^d \times \{0,1\}$ with the following interpretation: $x \in \mathbb{R}^d$ represents the features of an individual; the first feature $x_1$ 
(which we assume to be binary) represents a protected attribute (e.g., subgroup membership, black vs.~white) and we will also write it as $A \in \{0,1\}$; and $y \in \{0, 1\}$ represents the true label (e.g., ``re-offended'' or ``did not re-offend''). A labeled data set $S=(x^i,y^i)_{i=1}^n$ is a collection of such data points. We will partition $S$ into groups according to each individual's protected attribute and true label:
\begin{equation*}
S_{ay} = \{x^i \in S : x^i_1=a, y^i=y\}, ~a,y \in \{0,1\}
\end{equation*}
We write $\hat{Y}: \mathbb{R}^d\rightarrow \{0,1\}$ for a classifier that, given an individual's features, including her protected attribute, predicts her label.

Then, given a data set $S$ and a classifier $\hat{Y}$, writing $\hat{y}^i = \hat{Y}(x^i)$, we can formally define the false positive rate (FPR) and false negative rate (FNR) of $\hat{Y}$ on $S$ as follows:
\begin{align*}
FPR(\hat{Y}) &= \frac{\left|\{i:\hat{y}^i=1,y^i=0\}\right|}{\left|\{i: y^i=0\}\right|}\\
FNR(\hat{Y}) &= \frac{\left|\{i:\hat{y}^i=0,y^i=1\}\right|}{\left|\{i:y^i=1\}\right|}
\end{align*}

Given a value $a \in \{0,1\}$ of the protected attribute $A$, we denote by $FPR_{A = a}(\hat{Y})$, $FNR_{A = a}(\hat{Y})$ the false positive and false negative rates of $\hat{Y}$ on $\{(x,y) \in S : x_1 = a\}$.
\section{Penalizing Unfairness} \label{regularization}

Our approach to learning a fair classifier integrates fairness considerations into the learning process by penalizing unfairness, and is inspired by the concept of {\em regularization}. Typically in regularization, the added penalty term is a function only of the learned hypothesis, penalizing for complexity in the model, aiming to prevent overfitting. Here, we introduce a new type of penalty, which is not only hypothesis-dependent, but is also {\em data-dependent}, and which is set at a group level rather at an individual level. As our goal is to learn a classifier that matches FPR and FNR rates across populations, we define two types of penalizers that aim at minimizing the differences between the FPR and FNR (respectively) across sub-groups in the population, for the trained classifier.\footnote{Our penalization scheme minimizes the differences between the empirical FPR and FNR as evaluated on the relevant sub-groups in the training set $S$, and relies on statistical guarantees proven in Woodworth et al.~(\citeyear{woodworth}) to yield fairness on the true underlying distribution $\mathcal{D}$, for a sufficiently large dataset drawn i.i.d. from $\mathcal{D}$.}  




We focus our attention on boundary-based classifiers, which are trained in the form of a decision boundary in the feature space. In what follows we will assume the decision boundary is a hyperplane, and classification is therefore done in the following manner: Given a sample $x \in \mathbb{R}^d$, and a classifier $\hat{Y}$ specified by $\theta \in \mathbb{R}^d$, we predict $\hat{Y}(x) = sign(\theta^T x)$. We note that our approach also extends to the case of non-linear SVMs, by shifting to the space mapped into by the kernel function, and considering the problem of learning a decision hyperplane in that space. 

The first penalizer we propose is based on relaxing the 0-1 loss, to instead consider the margin from the decision boundary. We will penalize the difference in the average distance from the decision boundary across different values of  the protected attribute A.


We define the \textbf{Absolute Value Difference (AVD)} FPR penalty term to be
\begin{align*}
R_{FP}^{AVD}(\theta;S) &= \left|\dfrac{\sum\limits_{x\in S_{00}}^{} \theta^T x}{\left|S_{00}\right|} - \dfrac{\sum\limits_{x\in S_{10}}^{} \theta^T x}{\left|S_{10}\right|}\right| \\
&= \left|\theta^T\underbrace{\left(\dfrac{\sum\limits_{x\in S_{00}}^{} x}{\left|S_{00}\right|} - \dfrac{\sum\limits_{x\in S_{10}}^{} x}{\left|S_{10}\right|}\right)}_{\overline{x}}\right| \\
&= \left|\theta^T \overline{x} \right|
\end{align*}
The FNR penalty term is defined analogously. We note that this penalizer is convex in $\theta$. In order for the penalizer to also be differentiable at 0, we define a second variant (using the same notation for $\overline{x}$), which we term the \textbf{Squared Difference (SD)} penalizer:
\begin{equation*}
R_{FP}^{SD}(\theta;S) = \left(\theta^T \overline{x} \right)^2.
\end{equation*}
Again, we define the FNR penalizer analogously.
\begin{figure*}[ht]
  \includegraphics[width=1\textwidth]{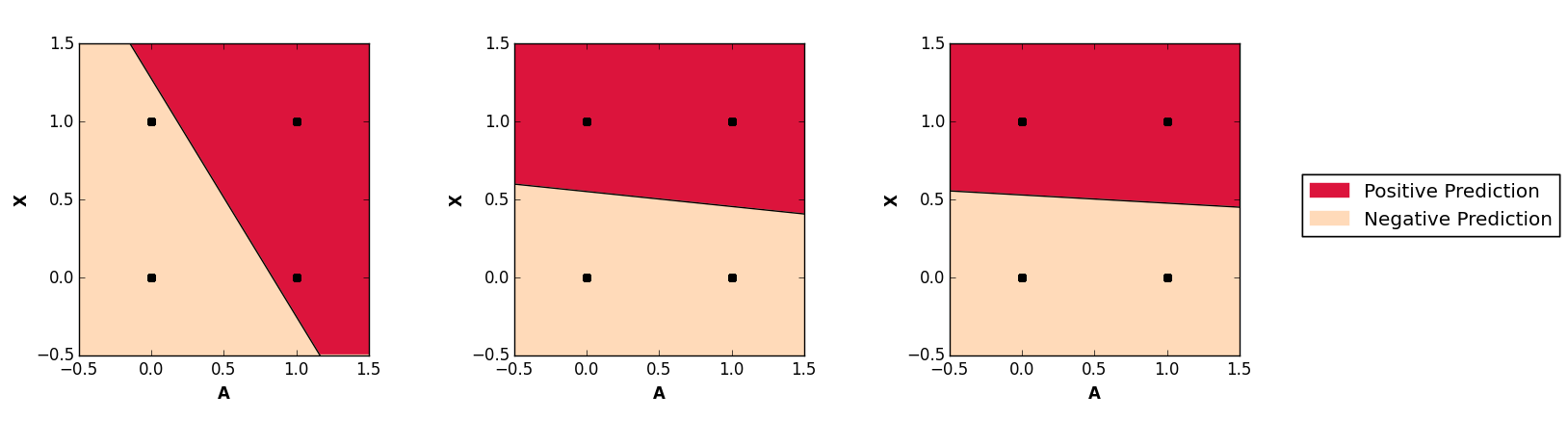}
  \caption{  Toy example with two attributes, $A$ (protected) and $X_2$ (non-protected). Classifying according to $A$ is perfectly unfair and has loss $\epsilon$. Classifying according to $X_2$ is perfectly fair and has loss $2 \epsilon$. Any post-processing of the optimal classifier (which classifies according to $A$ and loses the information contained in $X_2$) to ensure fairness will have the worst possible loss, $0.5$.  Here, we depict the halfspace learned via our penalization approach, increasing the weight assigned to the fairness penalizers (left-to-right). We see that placing sufficient weight on the penalizer results in prediction entirely according to $X_2$.
  This illustration was implemented based on random sampling 5,000 points from the distribution $\mathcal{D}_{\epsilon}$ described in Section~\ref{sec:incorporating}, with $\epsilon$ = 0.1, and learning using the scheme described in Section~\ref{sec:experiments}, using the $R_{FP}^{SD}$ and $R_{FN}^{SD}$ penalizers described in Section~\ref{regularization}, placing weights of $c_1=c_2=c$ for $c\in\{0,300,600\}$. }
  \label{fig:incorporating}
\end{figure*}

\section{The Importance of Incorporating Fairness in the Learning Phase} \label{sec:incorporating}
We briefly illustrate a simple example (based on one in Woodworth et al.~(\citeyear{woodworth})) which demonstrates the potential impact of incorporating fairness considerations into the learning process, rather than post-processing a learned classifier for fairness.

In the example, each data point 
lies in $X = (X_1, X_2) = \{0,1\}^2$ and has two features---$X_1=A$ is the protected attribute, and $X_2$ is a non-protected attribute---and a label in $Y = \{0,1\}$. Given $\epsilon \in (0,\frac{1}{4})$, we define a distribution $\mathcal{D}_{\epsilon}$ over labelled examples as follows:
\begin{align*}
\Prob[Y=1] &= 0.5 \\ 
\Prob[A=y|Y=y] &= 1-\epsilon \\
\Prob[X_2=y|Y=y] &= 1-2\epsilon
\end{align*}

Note that $\mathcal{D}_{\epsilon}$ is defined s.t. $A \perp X_2 | Y$.


\paragraph{Post-Processing}
Assume the hypothesis class $\mathcal{H}$ is unconstrained, and contains all of the (possibly randomized) functions $h:X \rightarrow \{0,1\}$. Note that classifying according to $X_2$ alone is a completely fair classifier (the FPR and FNR are both equal across $A = 0$ and $A=1$) that achieves 0-1 loss $2\epsilon$, and thus provides an upper bound on
\begin{equation*}
\min\limits_{h \in \mathcal{H}} \{L_{\mathcal{D}_{\epsilon}}^{0\textnormal{-}1}(h):h~\text{is perfectly fair}\}
\end{equation*}

The Bayes optimal predictor with respect to the 0-1 loss is
\begin{equation*}
\hat{h}(X) = \argmax\limits_{y \in \{0,1\}} \Prob[Y=1|X=x]
\end{equation*}
which, in our case, gives $\hat{h}(X) = A$. This classifier has 0-1 loss of only $\epsilon$. However, in terms of fairness, it performs as badly as possible, as it induces the maximal possible differences in both the FPR and FNR rates across the two sub-populations in the distribution.  

Any approach to post-processing this classifier for fairness (including, for example, the technique proposed by Hardt et al.~(\citeyear{hardt})) yields a classifier $\tilde{Y}$ that predicts 0 or 1 at random, each with probability 0.5. While $\tilde{Y}$ is a completely fair classifier, it only achieves trivial 0-1 loss of 0.5.

\paragraph{Incorporating Fairness in the Learning Process}
Now, assume that the hypothesis class $\mathcal{H}$ only contains (non-homogeneous) halfspaces $(w,b)$ over $\mathbb{R}^2$, and classification is done using the sign of $w^{T}X + b$, where $w \in \mathbb{R}^2$, and $b \in \mathbb{R}$.

Absent fairness considerations, the best separating halfspace (in terms of the 0-1 loss) would provide us with classifications identical (on the given distribution) to those of the Bayes optimal classifier, and thus would have 0-1 loss of $\epsilon$, while being maximally unfair. However, as shown in Figure~\ref{fig:incorporating}, using our proposed method of penalization (as described in detail in Section~\ref{casestudy}) yields a halfspace which induces equivalent performance on $\mathcal{D}_{\epsilon}$ as classifying according to $X_2$, resulting in 0-1 loss of $2\epsilon$ and perfect fairness.

\section{Case Study: Fair Classification Using Logistic Regression} \label{casestudy}
In this section, we instantiate our approach for achieving fairness, in the context of logistic regression. We denote the trained parameters of our logistic regressor by $\theta \in \mathbb{R}^d$, and the log-likelihood of $\theta$ given training set $S$ by $ll(\theta;S)$. 



We wish to solve the following optimization problem:

\begin{align}\label{eq:1}
\begin{split}
\underset{\theta}{\text{minimize}}
 ~~&L_{S}^{0\text{-}1}(\theta)\\
&+d_1|FPR_{A=0}(\theta;S)-FPR_{A=1}(\theta;S)|\\
&+d_2|FNR_{A=0}(\theta;S)-FNR_{A=1}(\theta;S)|
\end{split}
\end{align}
where $d_1, d_2 \geq 0$ are to be set up front, according to the desired trade-off between accuracy, FPR matching, and FNR matching. Applying our suggested relaxation ($R_{FP}$, $R_{FN}$ are to be set as either Absolute Value Difference or Squared Difference penalizers), and adding a standard $\ell_2$ regularization term, we get the following convex optimization problem:
\begin{align} \label{eq:2}
\begin{split}
\underset{\theta}{\text{minimize}}
~~&-ll(\theta;S) \\
&+c_1 R_{FP}(\theta;S) \\
&+c_2 R_{FN}(\theta;S)\\
&+q\left|\left|\theta\right|\right|_2^2 \\
\end{split}
\end{align}

For convenience, we will denote the objective in (\ref{eq:1}) by $\text{Objective}(\theta;S,d_1,d_2)$, and the objective in the proxy problem (\ref{eq:2}) by $\text{Proxy}(\theta;S,c_1,c_2,q)$. As the proxy is easy to solve using standard methods, we use it when optimizing, and then shift back to the original problem for estimating the quality of our results.



\section{Experiments}\label{sec:experiments}
We validate our approach using multiple datasets containing real-life data from the fields of criminal risk assessment, credit, lending, and college admissions. In each of the datasets we select a binary feature and treat it as the protected attribute (e.g., race or gender), which is the feature we require our trained classifier to behave fairly upon. Our proposed method performs well on all of these datasets, succeeding in removing unfairness almost entirely, at a very modest price in terms of accuracy.

\begin{table*}[h]
\centering
\resizebox{\textwidth}{!}{
\def\arraystretch{1.2}

\begin{tabular}{c c c | c | c | c || c | c | c || c | c | c |}

\cline{4-12}
&&&
\multicolumn{9}{ c| }{\textbf{COMPAS Dataset}}
\\ \cline{4-12}
&&&
\multicolumn{3}{ c|| }{\textbf{FPR Considerations}}&
\multicolumn{3}{ c|| }{\textbf{FNR Considerations}}&
\multicolumn{3}{ c| }{\textbf{Both Considerations}}
\\ \cline{4-12}
&&&
 $\mathbf{Acc.}$ &  $\mathbf{D_{FPR}}$ &  $\mathbf{D_{FNR}}$ &  $\mathbf{Acc.}$ &  $\mathbf{D_{FPR}}$ &  $\mathbf{D_{FNR}}$ &  $\mathbf{Acc.}$ &  $\mathbf{D_{FPR}}$ &  $\mathbf{D_{FNR}}$
\\  \cline{4-12}
\vspace*{-0.5ex}
\\ \cline{1-2} \cline{4-12}
\multicolumn{1}{ |c  }{} &
\multicolumn{1}{ c|  }{  \textbf{Our Method (AVD Penalizers)}}  &&
$\mathbf{0.660}$    &  $\mathbf{0.01}$  &  $0.04$ &
$\mathbf{0.653}$    &  $0.02$   &  $\mathbf{0.04}$ &
$\mathbf{0.654}$    &  $\mathbf{0.02}$  &  $\mathbf{0.04}$
\\ \cline{1-2} \cline{4-12}
\multicolumn{1}{ |c  }{} &
\multicolumn{1}{ c|  }{  \textbf{Our Method (SD Penalizers)}}  &&
$\mathbf{0.664}$    &  $\mathbf{0.02}$  &  $0.09$ &
$\mathbf{0.661}$    &  $0.05$   &  $\mathbf{0.03}$ &
$\mathbf{0.661}$    &  $\mathbf{0.02}$  &  $\mathbf{0.03}$
\\ \cline{1-2} \cline{4-12}
\multicolumn{1}{ |c  }{} &
\multicolumn{1}{ c|  }{  Zafar et al.~(\citeyear{disparatemistreatment})}  &&
$0.660$    &   $0.06$    &   $0.14$  &
$0.662$    &   $0.03$    &   $0.10$  &
$0.661$    &   $0.03$    &   $0.11$
\\ \cline{1-2} \cline{4-12}
\multicolumn{1}{ |c  }{} &
\multicolumn{1}{ c|  }{  Zafar et al. Baseline~(\citeyear{disparatemistreatment})}  &&
$0.643$    &   $0.03$    &   $0.11$  &
$0.660$    &   $0.00$    &   $0.07$  &
$0.660$    &   $0.01$    &   $0.09$
\\ \cline{1-2} \cline{4-12}
\multicolumn{1}{ |c  }{} &
\multicolumn{1}{ c|  }{  Hardt et al.~(\citeyear{hardt})}  &&
$0.659$    &  $0.02$    &   $0.08$  &
$0.653$    &  $0.06$   &    $0.01$  &
$0.645$    &  $0.01$   &    $0.01$
\\ \cline{1-2} \cline{4-12}
\multicolumn{1}{ |c  }{} &
\multicolumn{1}{ c|  }{  \textbf{Vanilla Regularized Logistic Regression}}  &&
$\mathbf{0.672}$    &   $\mathbf{0.20}$    &   $\mathbf{0.30}$  &
$\mathbf{0.672}$    &   $\mathbf{0.20}$    &   $\mathbf{0.30}$  &
$\mathbf{0.672}$    &   $\mathbf{0.20}$    &   $\mathbf{0.30}$
\\ \cline{1-2} \cline{4-12}
\end{tabular}
}
\vspace{3mm}
\caption{Performance comparison on the COMPAS dataset. For the approaches in bold -- Accuracy, FPR difference and FNR difference are evaluated on the test set, averaging over five runs and using a 70-30 training/test split. The performance of the remaining three approaches is stated as reported in Zafar et al.~(\citeyear{disparatemistreatment}).} \label{table:comparison_results}
\end{table*}

\begin{figure*}[b]
  \includegraphics[scale=0.6]{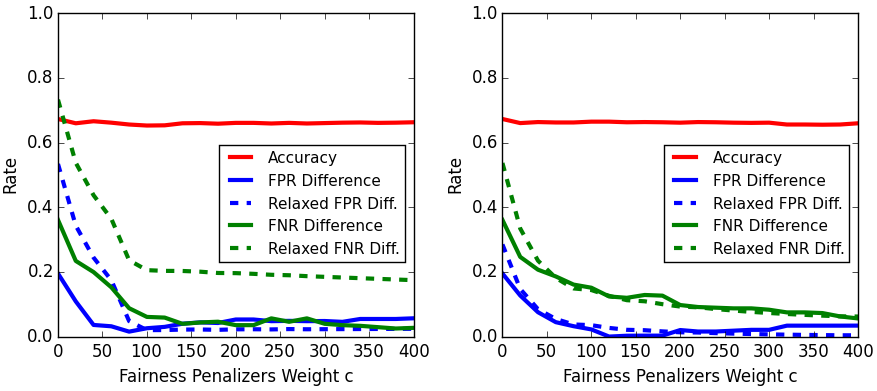}
  \caption{COMPAS Dataset. Accuracy, FPR difference ($\mathbf{D_{FPR}}$), and FNR difference ($\mathbf{D_{FNR}}$) (all evaluated on the test set) of the learned classifier, as a function of the weight $c=c_1 = c_2 \geq 0$ placed on the fairness penalizer terms. On the left we use the Absolute Value Difference (AVD) penalizer, and the Squared Difference (SD) penalizer on the right, both as presented in Section~\ref{regularization}. ``Relaxed FPR/FNR Diff.'' plots the value of the relevant penalization term.} 
  \label{fig:compas}
\end{figure*}

\subsection{Implementation}
\textbf{Our method} 
For the purpose of comparison with  Zafar et al.~(\citeyear{disparatemistreatment}) and Hardt et al.~\cite{hardt} on the COMPAS data, we use a parameter $c$ to induce three possible combinations of weights on the FPR and FNR penalization terms: $c = c_1$ and $c_2 = 0$; $c_1 = 0$ and $c = c_2$; and $c = c_1 = c_2$. For the other three datasets, we consider only $c = c_1 = c_2$.\footnote{The reason for varying the values of $c$ in the training phase is since we shifted to a proxy problem, in which we rely on the distance from the decision boundary rather the actual classifications. 
It is possible, of course, that even better results are attainable using our scheme with other combinations of $c_1, c_2$, and $q$.} To explore the accuracy/fairness trade-off curve for the relaxed optimization problem~(\ref{eq:2}), we train for different values of $c$, starting at $c=0$ (which is just standard logistic regression), and growing gradually.

Given a dataset $Q$ and fixing a $d_1, d_2 \in \{0, 1\}$ of interest, we use the following training scheme:
\begin{enumerate}
\item Split $Q$ at random into training set $S$ and test set $T$.
\item For each $c$, perform cross-validation on $S$ to select the corresponding best value $q_c$ for the regularization parameter.
\item For each $(c,q_c)$, let $\theta_c = \argmin\limits_{\theta} \text{Proxy}(\theta;S,c,c,q_c)$.
\item Select $\theta^* \in \argmin\limits_{\theta_c} \text{Objective}(\theta_c;S,d_1,d_2)$.
\item Evaluate performance using $\theta^*$ on test set $T$.
\end{enumerate}
We report the average of five such runs, each with a fresh training-test split.

We solve the relaxed convex optimization problem using the CVXPY solver. Due to stability issues with large training sets, we use a train/test split of 30-70 on the larger datasets, rather than 70-30 as on the COMPAS dataset\footnote{The code implementing our method can be found at https://github.com/jjgold012/lab-project-fairness}.

%
%
%



We briefly describe the other algorithmic approaches to which we compare:\\
\textbf{Zafar et al.}~(\citeyear{disparatemistreatment}) performs optimization by considering a proxy for the bias: the covariance between the samples' sensitive attributes and the signed distance between the feature vectors of misclassified users and the classifier decision boundary.\\
\textbf{Zafar et al. Baseline}~(\citeyear{disparatemistreatment}) tries to enforce equal FP/FN rates on the different groups by introducing different penalties for misclassified data points with different sensitive attribute values during the training phase.\\
\textbf{Hardt et al.}~(\citeyear{hardt}) performs post-processing on a standard trained (unfair) logistic regressor, picking different decision thresholds for different groups, and possibly adding randomization.

\subsection{Experimental Results}

In what follows, we use the following notation, given a trained classifier $\hat{Y}$:
\begin{align*}
\mathbf{D_{FPR}}&=\left|FPR_{A=0}(\hat{Y})-FPR_{A=1}(\hat{Y})\right| \\ 
\mathbf{D_{FNR}}&=\left|FNR_{A=0}(\hat{Y})-FNR_{A=1}(\hat{Y})\right|
\end{align*}
The values $FPR_{A=0}(\hat{Y})$, $FPR_{A=1}(\hat{Y})$, $FNR_{A=0}(\hat{Y})$, $FNR_{A=1}(\hat{Y})$ are reported as evaluated on the test set.

\paragraph{The COMPAS Dataset\footnote{https://github.com/propublica/compas-analysis}} The Correctional Offender Management Profiling for Alternative Sanctions (COMPAS) records from Broward County, Florida 2013-2014, made available online by ProPublica, are perhaps the best-studied data in the context of fairness.  The goal in this scenario is to successfully predict recidivism within two years, based on features such as age, gender, race, number of prior offenses, and charge degree. The dataset contains 5,278 samples. The protected attribute in this scenario is race, where $A$ indicates black or white. We filtered the dataset using the same features as Zafar et al.~(\citeyear{disparatemistreatment}), to allow for comparison.



\begin{table*}[t]
\centering
\caption{A description of the datasets used, along with parameters of the training procedure used for each.}
\label{table:datasets_description}
\begin{adjustbox}{max width=\textwidth}
\begin{tabular}{|l|l|l|l|l|l|l|l|}
\hline
\textbf{Dataset} & \textbf{No. Samples} & \textbf{No. Features} & \textbf{Train/Test Split} & \textbf{No. Repetitions} & \textbf{No. Folds in CV} & \textbf{Protected Feature} & \textbf{Target Variable} \\ \hline
COMPAS           & 5,278                     & 5                          & 70-30                     & 5                        & 5                                 & Race                       & 2-Year-Recidivism        \\ \hline
Adult            & 30,162                    & 10                         & 30-70                     & 5                        & 5                                 & Gender                     & Income Over/Under 50K    \\ \hline
Default          & 30,000                    & 23                         & 30-70                     & 5                        & 3                                 & Gender                     & Defaulting On Payments   \\ \hline
Admissions       & 20,839                    & 17                         & 30-70                     & 5                        & 3                                 & Race                       & Passing Bar Exam         \\ \hline
\end{tabular}
\end{adjustbox}
\end{table*}

\begin{table*}[t]
\centering
\resizebox{\textwidth}{!}{
\def\arraystretch{1.2}

\begin{tabular}{c c c | c | c | c || c | c | c || c | c | c |}

\cline{4-12}
&&&
\multicolumn{3}{ c|| }{\textbf{Adult Dataset}}&
\multicolumn{3}{ c|| }{\textbf{Default Dataset}}&
\multicolumn{3}{ c| }{\textbf{Admissions Dataset}}
\\ \cline{4-12}
&&&
 $\mathbf{Acc.}$ &  $\mathbf{D_{FPR}}$ &  $\mathbf{D_{FNR}}$ &  $\mathbf{Acc.}$ &  $\mathbf{D_{FPR}}$ &  $\mathbf{D_{FNR}}$ &  $\mathbf{Acc.}$ &  $\mathbf{D_{FPR}}$ &  $\mathbf{D_{FNR}}$
\\  \cline{4-12}
\vspace*{-0.5ex}
\\ \cline{1-2} \cline{4-12}
\multicolumn{1}{ |c  }{} &
\multicolumn{1}{ c|  }{  \textbf{Our Method (AVD Penalizers)}}  &&
$\mathbf{0.776}$    &  $\mathbf{0.00}$  &  $\mathbf{0.04}$ &
$\mathbf{0.807}$    &  $\mathbf{0.00}$   &  $\mathbf{0.01}$ &
$\mathbf{0.950}$    &  $\mathbf{0.01}$  &  $\mathbf{0.00}$
\\ \cline{1-2} \cline{4-12}
\multicolumn{1}{ |c  }{} &
\multicolumn{1}{ c|  }{  \textbf{Our Method (SD Penalizers)}}  &&
$\mathbf{0.783}$    &  $\mathbf{0.00}$  &  $\mathbf{0.09}$ &
$\mathbf{0.806}$    &  $\mathbf{0.01}$   &  $\mathbf{0.02}$ &
$\mathbf{0.950}$    &  $\mathbf{0.00}$  &  $\mathbf{0.00}$
\\ \cline{1-2} \cline{4-12}
\multicolumn{1}{ |c  }{} &
\multicolumn{1}{ c|  }{  \textbf{Vanilla Regularized Logistic Regression}}  &&
$\mathbf{0.800}$    &   $\mathbf{0.08}$    &   $\mathbf{0.39}$  &
$\mathbf{0.807}$    &   $\mathbf{0.01}$    &   $\mathbf{0.05}$  &
$\mathbf{0.951}$    &   $\mathbf{0.16}$    &   $\mathbf{0.02}$
\\ \cline{1-2} \cline{4-12}
\end{tabular}
}
\vspace{3mm}
\caption{Performance on the Adult, Loan Default, and Admissions datasets, penalizing for both FPR and FNR difference. Accuracy, FPR difference and FNR difference are evaluated on the test set, averaging over five runs and using a 30-70 training/test split.} \label{table:comparison_results_rest}
\end{table*}

In Table~\ref{table:comparison_results}, we compare the performance of our approach with that of three other techniques from the literature. Each method was trained based on logistic regression.  As a basis for comparison, we also present the performance of vanilla logistic regression, absent fairness considerations, with the regularization parameter selected via cross-validation.\footnote{Zafar et al.~(\citeyear{disparatemistreatment}) do not incorporate regularization in any of the approaches they report.}
Results for Zafar et al., Zafar et al. baseline, and Hardt et al. appear here as reported in Zafar et al.~(\citeyear{disparatemistreatment}).\footnote{Our method selects the classifier based on the training set only and reports its performance over the test set. Results for the three other approaches, reported by Zafar et al.~(\citeyear{disparatemistreatment}), are based on tuning parameters after seeing the trade-off curve over the test set, and reporting according to the best selection of these parameters.}

We find that the vanilla logistic regressor (absent fairness considerations) results in significant unfairness, as $\mathbf{D_{FPR}}=0.20$, and $\mathbf{D_{FNR}}=0.30$. The overall accuracy of this classifier measured on the test set was $0.672$.\footnote{Zafar et al.~(\citeyear{disparatemistreatment}) report a slightly different baseline of: Accuracy = 0.668, $\mathbf{D_{FPR}}=0.18$, $\mathbf{D_{FNR}}=0.30$.} Our SD penalization approach empirically achieves approximately the same accuracy as the Zafar et al.~(\citeyear{disparatemistreatment}) approach, with significantly better fairness. It is difficult to compare fairness-accuracy tradeoffs with the Hardt et al.~(\citeyear{hardt}) approach, since their accuracy is significantly lower than ours. A more direct comparison is possible by noting that our learned classifier can be post-processed to improve its fairness at a direct cost to accuracy. Hence, we can achieve accuracy of $0.659$ with $\mathbf{D_{FPR}} = \mathbf{D_{FNR}} = 0.01$, which compares very favorably with the Hardt et al. accuracy rate of 0.645 given the same FPR and FNR rates.\footnote{For completeness, we note that using a 50-50 training-test split (again not using the test set for parameter selection), our method (SD, both considerations) produces a classifier that provides: Accuracy = 0.659, $\mathbf{D_{FPR}} = 0.01, \mathbf{D_{FNR}} = 0.05$. This classifier can be post-processed to achieve rates of: Accuracy = 0.655, $\mathbf{D_{FPR}} = \mathbf{D_{FNR}} = 0.01$.}

Figure \ref{fig:compas} illustrates the accuracy/fairness trade-offs achievable using our scheme. Increasing the weight $c$ on the proxy fairness penalizers results in reducing their magnitude. The figure also illustrates how our relaxed penalizers succeed in tracking the real FPR and FNR differences. 

\subsection{Additional Datasets}

Table~\ref{table:datasets_description} provides summary statistics on each of the datasets on which we tested our approach. We also briefly describe the datasets below.

{\bf The Adult Dataset}\footnote{http://archive.ics.uci.edu/ml/datasets/Adult} is based on 1994 US Census data. The task we consider is to predict whether the income of each individual is over or under 50K dollars per year, based on features such as occupation, marital status, and education. The protected attribute selected in this task is gender. 

{\bf The Loan Default Dataset}\footnote{{\scriptsize https://archive.ics.uci.edu/ml/datasets/default+of+credit+card+clients}}
contains data regrading Taiwanese credit card users. The task we consider is to predict whether an individual will default on payments, based on features such as history of past payments, age, and the amount of given credit. The protected attribute is gender.

{\bf The Admissions Dataset}\footnote{http://www2.law.ucla.edu/sander/Systemic/Data.htm}
contains records of law school students who went on to take the bar exam. The task we consider is to predict whether a student will pass the exam based on features such as LSAT score, undergraduate GPA, and family income. The protected attribute is set to race.

\begin{figure*}[]
  \includegraphics[scale=0.6]{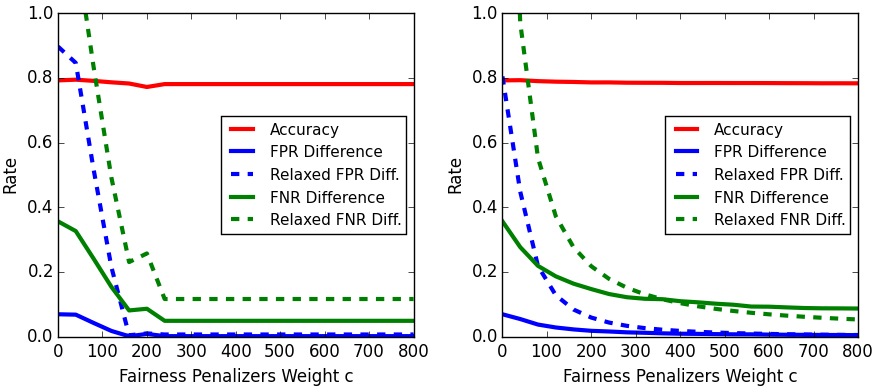}
  \caption{Adult Dataset. Fairness-Accuracy tradeoffs, as in Figure~\ref{fig:compas}.}
  \label{fig:adult}  
\end{figure*}

\begin{figure*}[]
  \includegraphics[scale=0.6]{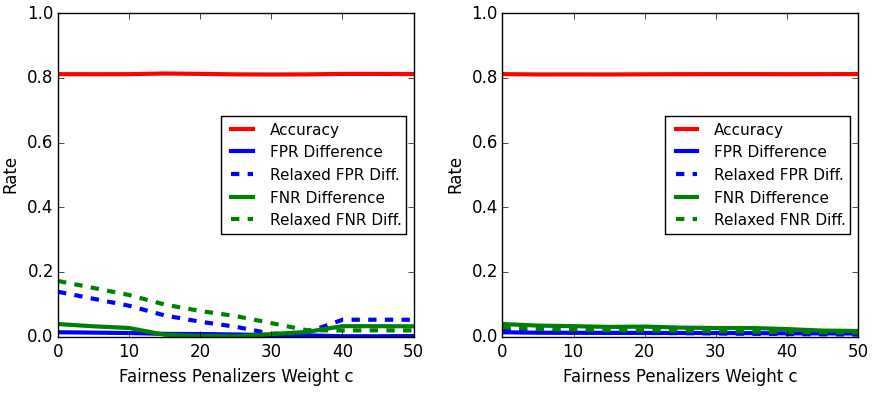}
  \caption{Loan Default Dataset. Fairness-Accuracy tradeoffs, as in Figure~\ref{fig:compas}.}
  \label{fig:default}
\end{figure*}

\begin{figure*}[]
  \includegraphics[scale=0.6]{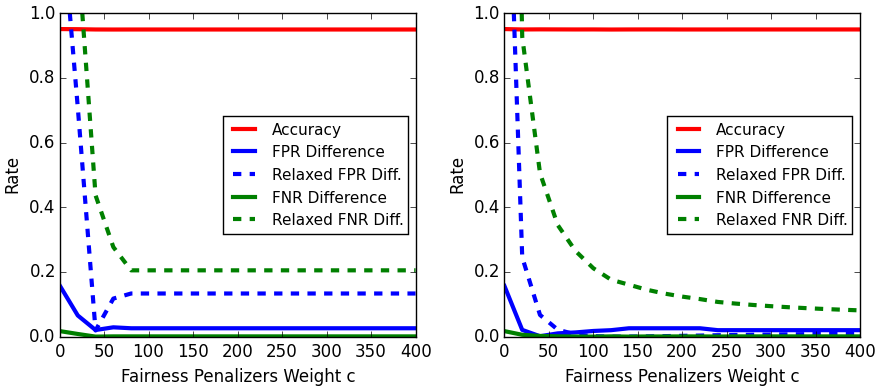}
  \caption{Admissions Dataset. Fairness-Accuracy tradeoffs, as in Figure~\ref{fig:compas}.}
  \label{fig:lawschool}
\end{figure*}

Table~\ref{table:comparison_results_rest} describes the performance of our approach on these datasets, and Figures~\ref{fig:adult},~\ref{fig:default}, and~\ref{fig:lawschool} illustrate the fairness-accuracy trade-offs we achieve in each context. Overall, we see that unfairness is nearly eliminated while accuracy remains quite high. The dataset on which accuracy suffers most under our approach is the Adult dataset, which is also the dataset on which the vanilla regression is the most unfair.

\section{Discussion}
Ensuring fairness in machine learning entails addressing the philosophical question of, given a particular setting, how fairness should formally be defined. Given a formal notion of fairness, the next question is how it can be achieved, and at what cost to accuracy. Over the past few years, FPR- and FNR-rate matching have emerged as compelling fairness notions deserving of attention. As we see from our experiments, fairness-unaware learning algorithms are sometimes extremely unfair according to these metrics. It is important, then, to ask what can be done to address this, and how accuracy will be impacted.

As learning optimal classifiers to match FPR and FNR across populations may be computationally intractable~\cite{woodworth}, it is natural that multiple approaches to this problem might emerge, each with its own pros and cons. Prior to our work, two groundbreaking papers had proposed approaches to ensuring FPR- and FNR-matching. Hardt et al.-style post-processing~\cite{hardt} is easy to implement and can be layered atop an already (unfairly) trained classifier. However, in some applications, because it does not integrate fairness in the learning process, it may be inherently sub-optimal. We illustrate this drawback in Section~\ref{sec:incorporating}. Some might also find post-processing distasteful, as it intentionally reduces accuracy on some individuals, in order to compensate for poor accuracy on others. The proxy-based approach of ~\cite{disparatemistreatment} makes nice use of the concept of disciplined convex-concave programming, however it has not been shown capable of lowering the unfairness below a certain (non-negligible) threshold.

As we show, fairness can successfully be achieved in many real-world settings via the addition of a judiciously chosen penalty term in the learning objective. We hope that this penalization approach, and the proxy we introduce for imposing FPR- and FNR-matching, will expand and enrich the toolkit for state-of-the art fair learning, and will help bring the goal of fair learning within reach.

\section{Acknowledgements}
This work was supported in part by NSF grants CNS-1254169 and CNS-1518941, US-Israel Binational Science Foundation grant 2012348, Israeli Science Foundation (ISF) grant \#1044/16, a subcontract on the DARPA Brandeis Project, and the HUJI Cyber Security Research Center in conjunction with the Israel National Cyber Bureau in the Prime Minister's Office.

\bibliography{citations}
\bibliographystyle{icml2018}

\end{document}